Oral (Poster is also fine)     Topic: *Deep Learning*

# Some like it tough: Improving model generalization via progressively increasing the training difficulty

**Hannes Fassold** [1]
[1] JOANNEUM RESEARCH – DIGITAL, Steyrergasse 17, 8010 Graz
E-mail: hannes.fassold@joanneum.at

**Summary:** In this work, we propose to progressively increase the training difficulty during learning a neural network model via a novel strategy which we call *mini-batch trimming*. This strategy makes sure that the optimizer puts its focus in the later training stages on the more difficult samples, which we identify as the ones with the highest loss in the current mini-batch. The strategy is very easy to integrate into an existing training pipeline and does not necessitate a change of the network model. Experiments on several image classification problems show that mini-batch trimming is able to increase the generalization ability (measured via final test error) of the trained model.

## 1. Introduction

Training a neural network model which generalizes well (has good performance on unseen data) is a highly desirable property, but is not easy to achieve. Nowadays, most often adaptive gradient methods like the *Adam* optimizer [1] are used for training a model as they are much easier to handle (less sensitive to weight initialization and hyperparameters) compared with mini-batch stochastic gradient descent (SGD). On the other hand, their generalization capability has been observed to be not as good as SGD [2].

In this work, we propose a simple strategy which we call *mini-batch trimming* to increase the generalization capability (measured for image classification problems as the error of the final model on the *test* dataset) of a trained model. The strategy is easy to integrate into an existing training pipeline, does not need a modification of the model structure and is independent of the employed optimizer (so can be used for both SGD and Adam-like methods). Its motivation lies from the fact that humans do learn subjects (e.g. algebra) 'from easy to hard': We first learn the basic concepts of a certain subject and learn the more advanced topics later. In the same way, we want our optimizer to focus in the later training stages on the more difficult samples in the dataset. E.g. for image classification, these are the ones which are harder to classify correctly.

Our strategy has similarities with *curriculum learning* methods and *importance sampling* methods. In curriculum learning (see the survey in [3]), during training the samples are presented in a more meaningful order (e.g. from easy to hard) instead of the default random order. *Importance sampling* methods do not treat all samples in a dataset in the same way, but instead bias the selection of samples via a certain criterion. E.g in [4], typicality sampling is used to overweight highly representative samples during training. A disadvantage of this approach is that it has a complicated workflow, which involves density clustering (via t-SNE algorithm [5]) in the sample space.

In the following section we will describe our proposed mini-batch trimming strategy, whereas in section 3 experiments will be done on standard image classification problems which demonstrate that the strategy leads to models which generalize better.

## 2. Mini-batch trimming

The training of a neural network model is usually done iteratively. In each iteration, a mini-batch consisting of *B* samples (where *B* is typically 64 or 128) is drawn randomly from the training set, the mean loss for the mini-batch is calculated in the forward pass and in the backward pass the gradient of the mean loss is utilized to update the model weights.

In order to focus more on the harder samples in the mini-batch, we propose a strategy which we call *mini-batch trimming*. As we cannot quantify the 'hardness' of a sample $\varphi$ exactly, we take the per-sample loss $L(\varphi)$ as an estimate of its hardness. This makes sense, as the more difficult samples in the training set typically also have a higher loss. We modify the forward pass now in the following way: First the per-sample loss $L(\varphi)$ is calculated for all samples in the mini-batch. Now all samples in the mini-batch are sorted using the per-sample loss as criterion. The mean loss is now calculated *only from a fixed fraction of the samples in the mini-batch with the highest per-sample loss*. So we are calculating sort of a *trimmed mean* instead of the usual mean. For selecting the fraction *p* of the samples with the highest loss the Pytorch framework provides the *torch.topk* operator, which is also differentiable.

In this way, in each training iteration the update of the model weights is biased towards the more difficult samples. In the fashion of curriculum learning, the fraction *p* is *linearly decreased* during training. For the first epoch *p* has the value 1.0 (take all samples in mini-batch into account), whereas in the last epoch *p* is set to 0.2 (focus only on the 20 % samples in the mini-batch with the highest loss). Experiments have shown that this is a sensible choice. Note that for neural networks without batch-normalization layers (e.g. transformer architectures for natural language processing), mini-batch trimming *brings also a runtime improvement*, as the backward pass then depends only on a part of the mini-batch [1].

## 3. Experiments and Evaluation

For the experiments and evaluation, we employ three standard datasets for image classification: SVHN, CIFAR-10 and CIFAR-100. The datasets consist of 32x32 pixel RGB images, which belong to either 10 classes (SVHN and CIFAR-10) or 100 classes (CIFAR-100). We use the Adam optimizer, with learning rate set to 0.001 and weight decay set to 0.0001. The mini-batch size is 128 and training is done for 150 epochs, with the learning rate decayed by a factor of 0.5 at epochs 50 and 100. We perform the experiments with two popular neural network architectures for computer vision, Resnet-34 [6] and Densenet-121 [7].

To measure how well the trained model is able to generalize, we utilize the top-1 classification error of the final model on the test set (which of course has not been seen during training). For each configuration, we do 10 different runs with random seeds and take the average of these 10 runs. We compare the standard training with the variant with mini-batch trimming enabled. Results of the experiments can be seen in Table 1. The evaluation shows that mini-batch trimming is able to improve the generalization capability of the model in nearly all cases, except for one case (Densenet-121 architecture on CIFAR-10 dataset) where there is a slight regression in the model performance.

## 3. Conclusion

We presented a novel strategy called mini-batch trimming for improving the generalization capability of a trained network model. It is easy to implement and add to a training pipeline and independent of the employed model and optimizer. Experiments show that the proposed method is able to improve the model performance in nearly all cases. In the future, we plan to investigate and integrate this strategy within a distributed training framework like DeepSpeed.

*Table 1: Comparison of training with mini-batch trimming disabled / enabled for various network architectures and datasets. The first value in each cell is the average test error (in percent, averaged over 10 runs) with mini-batch trimming disabled, the second value is with mini-batch trimming enabled. The lower value is marked in bold.*

| Dataset | Network architecture | |
|---|---|---|
| | Resnet-34 | Densenet-121 |
| SVHN | 5.87 / **5.76** | 4.62 / **4.42** |
| CIFAR-10 | 17.43 / **17.01** | **10.10** / 10.19 |
| CIFAR-100 | 48.19 / **47.72** | 32.95 / **32.18** |

## Acknowledgements

The research leading to these results has received funding from the European Union's Horizon 2020 research and innovation programme under grant agreement No. 951911 - AI4Media.

## References

[1]. D. Kingma, J. Ba, Adam: A Method for Stochastic Optimization, in *International Conference for Learning Representations (ICLR)*, 2015.
[2]. P. Zhou, J. Feng, C. Ma, C. Xiong, S. Hoi, E. Weinan, Towards theoretically understanding why SGD generalizes better than ADAM in Deep Learning, in *Conference on Neural Information Processing Systems (NeurIPS)*, 2020.
[3]. X. Wang, Y. Chen, W. Zhu, A Survey on Curriculum Learning, in *IEEE Transactions on Pattern Analysis and Machine Intelligence (TPAMI)*, 2021.
[4]. X. Peng, L. Li, F. Wang, Accelerating Minibatch Stochastic Gradient Descent using Typicality Sampling, in *IEEE Transactions on Neural Networks and Learning Systems*, 2020.
[5]. L. van der Maaten, G. Hinton, Visualizing Data using t-SNE, in *Journal of Machine Learning Research*, 2008.
[6]. K. He, X. Zhang, S. Ren, J. Sun, Deep Residual Learning for Image Recognition, in *IEEE Conference on Computer Vision and Pattern Recognition (CVPR)*, 2016.
[7]. G. Huang, Z. Liu, L. van der Maaten, K. Weinberg, Densely Connected Convolutional Networks, in *IEEE Conference on Computer Vision and Pattern Recognition (CVPR)*, 2017.

---

[1] https://stackoverflow.com/questions/68920059/pytorch-no-speedup-when-doing-backward-pass-only-for-a-part-of-the-samples-in-m